\PassOptionsToPackage{unicode}{hyperref}
\PassOptionsToPackage{hyphens}{url}
\documentclass[]{article}

\usepackage{authblk} 
\author[1]{Stéphanie M. van den Berg}
\author[2]{Ulrich Halekoh}
\author[2]{Sören Möller}
\author[3]{Andreas Kryger Jensen}
\author[2]{Jacob von Bornemann Hjelmborg}
\affil[1]{University of Twente}
\affil[2]{University of Southern Denmark}
\affil[3]{University of Copenhagen}

\usepackage{xcolor}
\usepackage[margin=1in]{geometry}
\usepackage{amsmath,amssymb}
\usepackage{iftex}
\ifPDFTeX
  \usepackage[T1]{fontenc}
  \usepackage[utf8]{inputenc}
  \usepackage{textcomp} 
\else 
  \usepackage{unicode-math} 
  \defaultfontfeatures{Scale=MatchLowercase}
  \defaultfontfeatures[\rmfamily]{Ligatures=TeX,Scale=1}
\fi
\usepackage{lmodern}
\ifPDFTeX\else
\fi
\IfFileExists{upquote.sty}{\usepackage{upquote}}{}
\IfFileExists{microtype.sty}{
  \usepackage[]{microtype}
  \UseMicrotypeSet[protrusion]{basicmath} 
}{}
\makeatletter
\@ifundefined{KOMAClassName}{
  \IfFileExists{parskip.sty}{%
    \usepackage{parskip}
  }{
    \setlength{\parindent}{0pt}
    \setlength{\parskip}{6pt plus 2pt minus 1pt}}
}{
  \KOMAoptions{parskip=half}}
\makeatother
\usepackage{longtable,booktabs,array}
\usepackage{calc} 
\usepackage{etoolbox}
\makeatletter
\patchcmd\longtable{\par}{\if@noskipsec\mbox{}\fi\par}{}{}
\makeatother
\IfFileExists{footnotehyper.sty}{\usepackage{footnotehyper}}{\usepackage{footnote}}
\makesavenoteenv{longtable}
\usepackage{graphicx}
\makeatletter
\newsavebox\pandoc@box
\newcommand*\pandocbounded[1]{
  \sbox\pandoc@box{#1}%
  \Gscale@div\@tempa{\textheight}{\dimexpr\ht\pandoc@box+\dp\pandoc@box\relax}%
  \Gscale@div\@tempb{\linewidth}{\wd\pandoc@box}%
  \ifdim\@tempb\p@<\@tempa\p@\let\@tempa\@tempb\fi
  \ifdim\@tempa\p@<\p@\scalebox{\@tempa}{\usebox\pandoc@box}%
  \else\usebox{\pandoc@box}%
  \fi%
}
\def\fps@figure{htbp}
\makeatother
\providecommand{\keywords}[1]
{
  \small	
  \textbf{\textit{Keywords---}} #1
}

\newcommand{\nn}{\nonumber}
\newcommand{\E}{\mathbb{E}}
\makeatletter
\@ifundefined{pandocbounded}{\newcommand{\pandocbounded}[2]{#2}}{}
\makeatother
\usepackage{mathtools, amsmath, amssymb, tikz, proofread}
\usepackage{bookmark}
\IfFileExists{xurl.sty}{\usepackage{xurl}}{} 
\hypersetup{
  pdftitle={Towards a pretrained deep learning estimator of the Linfoot informational correlation},
  pdfauthor={Stéphanie M. van den Berg, Ulrich Halekoh, Sören Möller, Andreas Kryger Jensen and Jacob von Bornemann Hjelmborg},
  hidelinks,
  pdfcreator={LaTeX via pandoc}}

\title{Towards a pretrained deep learning estimator of the Linfoot informational correlation}

\date{}

\usepackage{amsthm}

\newtheorem{proposition}{Proposition}[section]

\theoremstyle{definition}

\theoremstyle{definition}

\theoremstyle{definition}

\theoremstyle{definition}

\theoremstyle{remark}

\begin{document}
\maketitle

\usetikzlibrary{arrows,decorations.pathmorphing,decorations.pathreplacing,backgrounds,positioning,fit,matrix}

\abstract{
We develop a supervised deep-learning approach to estimate mutual information between two continuous random variables. As labels, we use the Linfoot informational correlation, a transformation of mutual information that has many important properties. Our method is based on ground truth labels for Gaussian and Clayton copulas. We compare our method with estimators based on kernel density, $k$-nearest neighbours and neural estimators. We show generally lower bias and lower variance. As a proof of principle, future research could look into training the model with a more diverse set of examples from other copulas for which ground truth labels are available. 
}

\keywords{mutual information, supervised learning, convolutional neural network, Clayton copula}

\section{Introduction}\label{introduction}

Dependency in bivariate data can have many forms. Most measures focus on monotone relationships, where one variable increases monotonically with an increase of another variable. The most well-known measure for (linear) dependency is the Pearson product-moment correlation, that is suitable for quantifying the dependency in bivariate Gaussian random variables.

Another way to look at dependency is from an information-theoretic point of view. This offers a general, model-independent way of quantifying dependency. For example, it also allows for non-monotone relationships. Mutual information (equivalently, the Kullback-Leibler divergence) quantifies dependency in terms of how much information is shared between two continuous random variables. Whereas a Pearson correlation is bounded by -1 and 1, mutual information is bounded by 0 and plus infinity, which makes it harder to interpret.

The Linfoot informational correlation \cite{Linfoot1957} is a transformation of the mutual information and satisfies the very much desirable properties stated by \cite{Renyi1959}, the most notable being equal to 0 under strict independence and being equal to 1 if there is a fully functional relationship. Furthermore, it is invariant to strictly monotone or injective transformation, and in case of bivariate Gaussian random variables, the Linfoot informational correlation agrees with the (absolute value of) the Pearson product-moment correlation.

For non-Gaussian data, the Linfoot correlation excels in providing insights to dependencies not obtainable by the Pearson linear correlation measure. Highly dependent random variables might have Pearson correlation of zero, while the Linfoot correlation will be positive due to the Rényi properties. Take the example of the fully functional relationship between \(X\) and \(Y = X^2\): a Pearson correlation will be 0 (given a marginal distribution of $X$ that is uniform on $[-1,1]$), whereas the Linfoot correlation will be 1, acknowledging the functional relationship.

The Linfoot informational correlation, as is mutual information, is in general not straightforwardly estimable. Several proposals for mutual information estimators have been suggested over the last decades and their performance has been evaluated in a variety of scenarios (\cite{Blumentritt2012, AlizadRahvar2011,Kraskov2004, Evans2008, pmlr-v80-belghazi18a, hu2024infonet, beyond-normal-2023, pmi-profiles-2025, Loftsgaarden1965}, to list just a few. In general, bias is hard to control in this task and several alternative measures to gauge dependency have been suggested by for example \cite{reshef2011detecting} and \cite{murrell2016discovering}.

Neural estimators of mutual information, discussed below, also show unwanted behaviour in the sense that their variance explodes with increasing dependence \cite{shin2024experimental}. It therefore makes sense to instead focus on estimating the Linfoot informational correlation, where a value of 1 is the limit as the mutual information goes to infinity.

We emphasize that it is hard to obtain an estimator that complies to the Rényi properties or approximates these in a satisfactory manner. In particular the transformation invariance property needs careful consideration. For instance, one may aim for invariance to a restricted class of injective trans formations preserving also continuity (see \cite[Thm. 2.1]{beyond-normal-2023}). We do not strive here for an overall general estimator that can handle any possible joint density. We aim for an estimator general enough to handle smooth densities, for instance without any spikes or holes. We focus on applications where pairs of random variables that share marginal and joint densities that are non-Gaussian should not be `far from being Gaussian' in terms of distributional and topological invariance properties of the data.

In many applications, prior to quantifying dependency with a Pearson correlation, the marginals are transformed to approximate normality using for instance a Box-Cox transformation. Of course, marginal normality does not imply multivariate normality, so this is a suboptimal approach. Depending on the transformation, the Pearson correlation has a different value. This is so since any nonlinear transformation changes the linear dependency. We propose instead to apply our estimator of the scale-invariant Linfoot informational correlation. The Linfoot correlation is invariant to nonlinear transformation of the marginals. Of course, due to finite sampling, an estimator may yield different values, but the estimand is the same.

We first describe several existing estimators of mutual information that can be used when the true joint density is unknown (nonparametric and neural estimators) and compare their performance in estimating the Linfoot informational correlation. We do that in the context of two archimedian copulas: the Gaussian copula and the Clayton copula. The former is parametrized by the Pearson correlation, which is equal to the Linfoot informational correlation. For the Clayton copula, we present a closed formula for the Linfoot correlation in terms of the Clayton dependence parameter, which is to our knowledge novel. Next, we propose a data-driven approach based on a deep learning neural network (DNN) and compare its performance to existing estimators. The application of neural networks to mutual information estimation has been adopted in for example \cite[\cite{hu2024infonet}]{pmlr-v80-belghazi18a}, known as neural estimators. There, neural networks are trained and tested on the same data sets. Heavily overparametrized DNN models are capable of representing complex structured joint distributions and are therefore appealing for the estimation task. Here we focus on training a supervised DNN: supervised learning of bivariate joint distributions that are associated with given Linfoot values.

In Section 2 we describe the relationship between the Clayton copula dependence parameter and the Linfoot correlation. This result is used to identify the Clayton copula parameter with
the Linfoot informational correlation.

In Section 3, we evaluate the performance of several existing estimators of mutual information that can be used to estimate the Linfoot informational correlation. We explore data conforming to both Gaussian and Clayton copulas, with various degrees of dependency and various sample sizes. From a given Linfoot informational correlation we simulated data with Gaussian marginals and either a Gaussian copula or a Clayton copula.

Section 4 proposes a data-driven approach based on deep learning. It describes our methodology to train deep learning neural network models that can estimate a Linfoot informational correlation. We compare performance to the estimators discussed in Section 3.

In Section 5 we apply our supervised deep learning Linfoot estimator to twin data on body mass index (BMI) and assess whether the information shared between identical twins is larger than the information shared between non-identical twins, which can be regarded as evidence for genetic inheritance.

In Section 6 we discuss how our supervised deep learning approach could be generalized further to other bivariate distributions.

\section{Clayton copula parameter's link to the Linfoot correlation}\label{Clayton}

For evaluating a Linfoot informational correlation estimator, it is necessary to know the ground truth.

The Linfoot informational correlation is a function of the mutual information:

\begin{align}
  L(u, v) = \sqrt{1- \exp(-2I(u;v))}
\end{align}

For a bivariate Gaussian distribution, the mutual information has the functional relationship
\(I(u;v) = - \frac{1}{2}\log(1- \rho^2)\)
with the Pearson correlation \(\rho\)
and the Linfoot informational correlation is equal to the Pearson correlation.

Estimating mutual information for an empirical data set that is bivariate Gaussian has been shown to be easier than for non-Gaussian distributions \cite{beyond-normal-2023}. As a special alternative case we consider the Clayton copula for which we derive a closed-form expression for the Linfoot information correlation in terms of the copula parameter, to obtain a ground truth for the evaluation.

The Clayton copula is a member of the family of Archimedean copulae where the copula function can be decomposed as
\begin{align}
  C(u,v) = \psi(\phi(u)+\phi(v))
\end{align}
for a continuous, strictly decreasing and convex function \(\phi: [0,1] \to [0,\infty)\) with \(\phi(1)=0\) and pseudo-inverse \(\psi\)
\cite{Nelsen2006}. In Proposition \ref{prop:ClaytonLinfoot} we present an expression of the Linfoot informational correlation for any Archimedean copula given by a functional of the inverse generator. The Clayton copula has inverse generator equal to \(\psi_\theta(t) = (1+\theta t)^{-1/\theta}\) for \(\theta \in (-1,\infty)\setminus\{0\}\) and using the general result for Archimedean copulae, a closed-form expression of the Linfoot informational correlation for Archimedean copulae is stated.

\begin{proposition}
Let $(X,Y)$ be a pair of continuous random variables with copula density function 
$c\colon\, [0;1]^2 \mapsto \mathbb{R}^+$. The Linfoot informational correlation between $X$ and $Y$ is given by
\begin{align}
  L(X,Y) &= \sqrt{1 - \exp\left(-2\iint\limits_0^1 c(u,v) \log c(u,v) du dv\right)}\label{eq:linfootCopula}
\end{align}
\label{prop:linfootCopula}
\end{proposition}
\begin{proposition}
Let the dependence between $X$ and $Y$ be given by an Archimedean copula with inverse generator $\psi$. The Linfoot informational correlation is then
\begin{align}
L_\theta(X,Y) &= \sqrt{1-\exp\left(-2 \iint\limits_0^\infty \psi_\theta^{\prime\prime}\left(x+y\right) \log \left(\frac{\psi_\theta^{\prime\prime}\left(x+y\right)}{\psi_\theta^{\prime}\left(x\right)\psi_\theta^{\prime}\left(y\right)} \right) dxdy\right)}\label{eq:linfootGenerator}
\end{align}
where $\psi^\prime_\theta$ and $\psi^{\prime\prime}_\theta$ are the first and second derivatives of the inverse of the generator.

Considering the special case of a Clayton copula with parameter $\theta > 0$, the Linfoot informational correlation between $X$ and $Y$ is
\begin{align}
L_\theta(X,Y) &= \sqrt{1-\exp\left(-2\left(\log(1+\theta)  - (2\theta+1)\left(1+\frac{1}{1+\theta}\right) + 2(1+\theta)\right)\right)}\label{eq:claytonLinfoot}
\end{align}
\label{prop:ClaytonLinfoot}
\end{proposition}

Both propositions are proved in the appendix.
Proposition \ref{prop:ClaytonLinfoot} also holds for the independence copula as this corresponds to the limit of \(\theta \searrow 0\) where \(\lim_{\theta \searrow 0} \psi_\theta(t) = \exp(-t)\) corresponding to the inverse generator of the independence copula. It is seen directly from equation \eqref{eq:linfootGenerator} that the Linfoot informational correlation is equal to zero in the limit.

\section{Comparing several existing estimators for the Linfoot correlation}\label{nonparam}

For Gaussian and Clayton copulas, we evaluated the performance of several common estimators for mutual information. We calculated five non-parametric estimators, of which three are based on a k-nearest neighbour(KNN) approach and two on kernel density estimation (KDE).
Besides the nearest neighbour estimator proposed by \cite{Loftsgaarden1965}, (KNN), we considered a truncated version that truncates a bin by the area that lies outside the boundaries of the unit square \cite{AlizadRahvar2011,Blumentritt2012}, (KNN truncated). Additionally, a nearest neighbour estimator was used that targets the mutual information via the digamma function instead of the copula density \cite{Evans2008,Kraskov2004}, (FNN).
The supremum norm was chosen for the determination of neighbouring points (KNN: \(n^{\frac{3}{7}}\), KNN truncated: \(n^{\frac{2}{3}}\), FNN: \(n^{\frac{3}{5}}\)). 

For the kernel density estimation of the copula density we corrected for boundary bias either by using the mirror reflection technique (KDE MR) \cite{Charpentier2006} or Beta kernels \cite{Blumentritt2012}, (KDE MR, bandwidth \(n^{-\frac{1}{4}}\), KDE beta, bandwidth \(n^{-\frac{3}{7}}\)).

The KNNs were obtained by straightforward implementation in R, the \texttt{FNN} by the R package \texttt{FNN} \cite{RFNN}.
Both the mirror reflected (MR) kernel density estimator and the estimator based on the Beta kernel are implemented in the R package \texttt{kdecopula} \cite{Rkdecopula}.

A special class of estimators is formed by neural estimators. Their purpose is estimating or maximising mutual information. By simple transformation, we get estimates for the Linfoot informational correlation. There are several types of neural estimators, but they all have in common that a relatively simple neural network, say a two-layer perceptron, is trained on the data set for which we would like to obtain an estimate. In discriminative estimators, the model is fed with examples from the true distribution, say a bivariate Gaussian with correlation \(\rho\), and learns to discriminate these examples from examples from the distribution that shows independence, achieved by permuting the values of one of the two marginals. One essentially builds a classifier that discriminates between dependent and independent examples. In generative estimators, the conditional distribution or the joint distribution are modelled first for example by a variational autoencoder or a generative adversarial network (GAN), respectively, after which the mutual information is computed. Both types have their pros and cons, but what they share is that the data set is used for training a network that is used to estimate the dependency in that same data set.

Neural estimators are often applied in machine learning, where they can be used for example to identify the highest mutual information for features downstream in a larger (deep) neural network. The few studies that evaluated their performance as an estimator are limited. They show that they generally  work better for Gaussian than non-Gaussian data, that their bias can increase exponentially with increasing mutual information, and they show high variance \cite{shin2024experimental}. Because they are dependent on training a network and depend on randomly reshuffling the data (the discriminative ones), they even show large variance on the \emph{same} dataset. Also, because the data is used both for training and for estimating the mutual information conditional on the trained network, they require large sample sizes.

We discuss neural estimators here, because they share some similarities with our newly proposed estimator discussed in the next section. We report a simple implementation of the MINE estimator \cite{belghazi2018}, using a two-layer perceptron of 64 and 32 nodes respectively, as implemented in the \texttt{torch\_mist} Python package \cite{torchmist2025}.

We tested the estimators on simulated data with known Linfoot correlation. We varied sample size (100, 200, 500, 1000) and the strength of the dependence in terms of the Linfoot correlation (0, .2, .4, .6, .8, .99). We simulated data from Gaussian and Clayton copulas. We simulated the data such that the marginals were Gaussian. For each combination of copula type, Linfoot correlation, and sample size, we simulated 1000 data sets. The results are in Tables 1 and 2 (Appendix). There were convergence problems for the KNN estimators for sample sizes of 1000. Overall best performance was by the FNN estimator. We will use the FNN to benchmark the performance our own proposed estimator.

\section{Pre-trained deep neural network}\label{pre-trained-deep-neural-network}

We propose a pretrained deep neural network that can be used as a fast estimator of the Linfoot informational correlation. A purely data driven approach was taken, by having a neural network learn the relation between data and the Linfoot correlation by presenting it with numerous examples. Many data sets were simulated using various Linfoot correlations, different sample sizes, and different distributions.

There were three different types of neural network models that were trained on the examples. A first model was trained on handcrafted features (summary statistics). A second model was trained by using the raw data represented as heatmaps. The two-dimensional heatmap images were then given as input to a deep neural network that included several convolutional layers. A third model consisted of a combination of these two approaches using feature fusion based on vector concatenation (Gong Cheng, Junwei Han, 2016).

The target values for the supervised learning by the neural network were the Linfoot correlations used for generating the data sets.

\subsection{Model 1}\label{model-1}

The architecture
of the first model `Model 1' using only 56 sample statistics as input features was relatively simple, consisting of three dense layers of 32 nodes with ReLu activation functions, a pruning layer that dropped 25\% of the parameters to zero, and a final one-node output layer.

\subsection{Model 2}\label{model-2}

The images were input for a convolutional neural network, with ReLu activation functions. The layers consisted of a rescaling
layer, two convolutional layers (16 features, kernel 9x9), a batch normalisation layer,
a maxpooling layer (2x2), a dropout (25\%), another convolutional layer (16 features, kernel 3x3), and a max pooling layer (2x2). After flattening, there is another
dropout of 50\%, then a dense layer of 32 nodes, another dropout of 25\%, a dense
layer of 32 nodes, and a single-node output layer. 

\subsection{Model 3}\label{model-3}

Model 3, the combined model had as input both the images and the 56 features. For the architecture we used Model 2 for the image data and Model 1 for the handcrafted features. The output features from the CNN were then combined with
the output features of the simple model for the 56 input features using vector concatenation and fed into an additional neural network with four dense layers, again with ReLu activation functions.

The models were trained to make predictions for the Linfoot correlations. If a network model
predicted a Linfoot correlation less than 0, it was set to 0. If a network model
predicted a Linfoot correlation of more than 1, it was set to 1.

\subsection{Model training and testing}\label{model-training-and-testing}

Training data were simulated from Linfoot correlations between 0
and 0.99, in such a way that 90\% of the correlations were
uniformly distributed between 0 and 0.99, and 10\% were
uniformly distributed between 0 and 0.01. This oversampling was done because earlier runs with the pretrained estimator showed difficulty with estimating correlations close to 0. For each randomly sampled
Linfoot correlation, data were generated following a bivariate Gaussian
copula or a Clayton copula.

Different sample sizes were used: 100, 200, 500, and 1000.
For each sample size, there were
4000 datasets of Gaussian distributed data and 4000 datasets of Clayton
distributed data. The marginals were simulated to be Gaussian.

For each of the
4 (samples sizes) \(\times\) 2 (distributions) \(\times\) 4000 (replications) = 32000 training datasets, 56 features were
computed (sample statistics). First, the FNN estimator was computed with tuning parameter \(k\) equal to \(n^\frac{3}{5}\). Second, the Pearson product-moment estimator was computed. For the remaining 54 features, the range of values (from minimum to maximum observed value per dataset) of one variable was divided into 18 bins. For each bin, the mean, the variance
and the skewness were calculated for the other variable. Thus, for 18 marginal
distributions, the first three moments were computed. The number of 18 was chosen to have a reasonable
number of observations per bin to compute variance and skewness, even for small sample sizes. In case there
were too few observations to compute variance and/or skewness, the statistic was fixed to 0.
Together with the FNN and Pearson
estimators, these formed \(2+ 18 \times 3 = 56\) handcrafted features to be fed
to a neural network to predict the true Linfoot correlations.

For the image input for Models 2 and 3, both variables were binned into 50 intervals of equal length, creating \(50 \times
50 = 2500\) pixels in a \(50 \times 50\) grid. For each pixel, the number of observed datapoints was counted and standardised by dividing by the total number of data points. The result is then effectively a grey-scale heatmap.

Target values consisted of the Linfoot correlations
used for generating the training data. The loss function was the mean squared error and an Adam optimiser was used with learning rate 0.001 with mean squared error as the loss function. During the
optimisation consisting of 50 epochs, batch size 1000, the predictions were validated on independent validation data: 500 data sets generated for each combination of copula, sample size and Linfoot correlation used in the test data.

All estimators (Model 1, Model 2, Model 3 and the FNN as benchmark) were tested on test data sets with the same structure that was
used for the comparisons of the non-parametric and neural estimators in Sec. \ref{nonparam}.

\subsection{Results}\label{results}

The result of the the evaluation of the existing estimators are in Tables 1 and 2. The estimator with overall lowest bias and lowest variance was the FNN, and this one was used to benchmark the performance of our own supervised deep neural network models.

Of the three supervised deep neural network models, Model 1 performed better than Model 2, but both were outperformed by Model 3 that combined handcrafted features with a heatmap image. We only present the comparison of Model 3 (``pretrained'') with the FNN benchmark. Figure 1 presents the bias of the estimators. The bias gets smaller with increasing sample size, and the bias is smaller for the Gaussian copula than for the Clayton copula. The mostly negative bias is smallest for correlations of 0.99. For zero correlations, the bias is strongly positive, because estimates can only have positive values. The pretrained neural network estimator and the FNN estimator always yielded valid estimates. The pretrained neural network estimator shows less bias than the FNN estimator. 
\begin{figure}[h]
\caption{Estimation bias for various sample sizes and different copula.}
\includegraphics[width=0.9\textwidth]{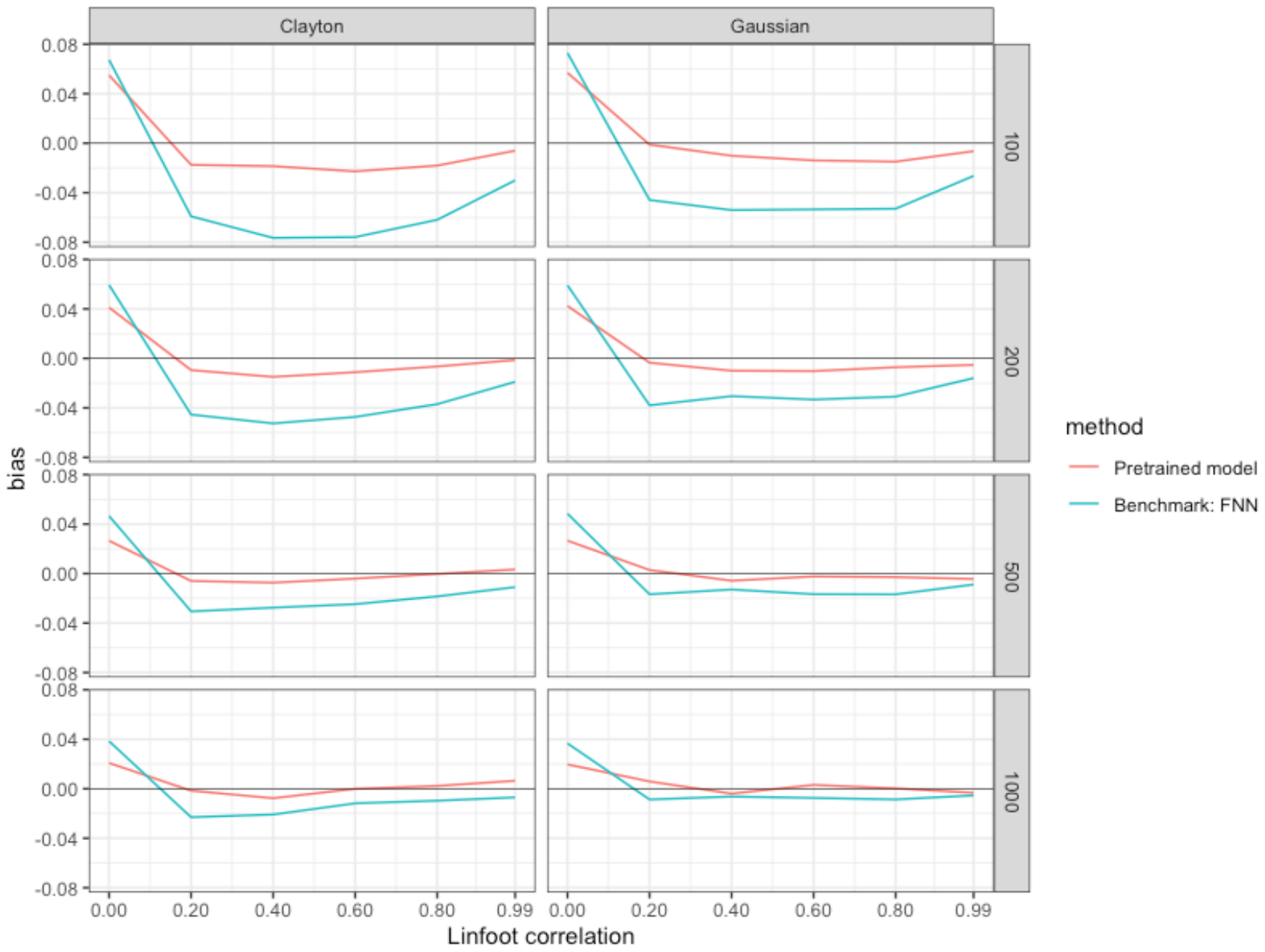}
\end{figure}

\begin{figure}[h]
\caption{Estimation standard deviation for various sample sizes and different copula.}
\includegraphics[width=0.9\textwidth]{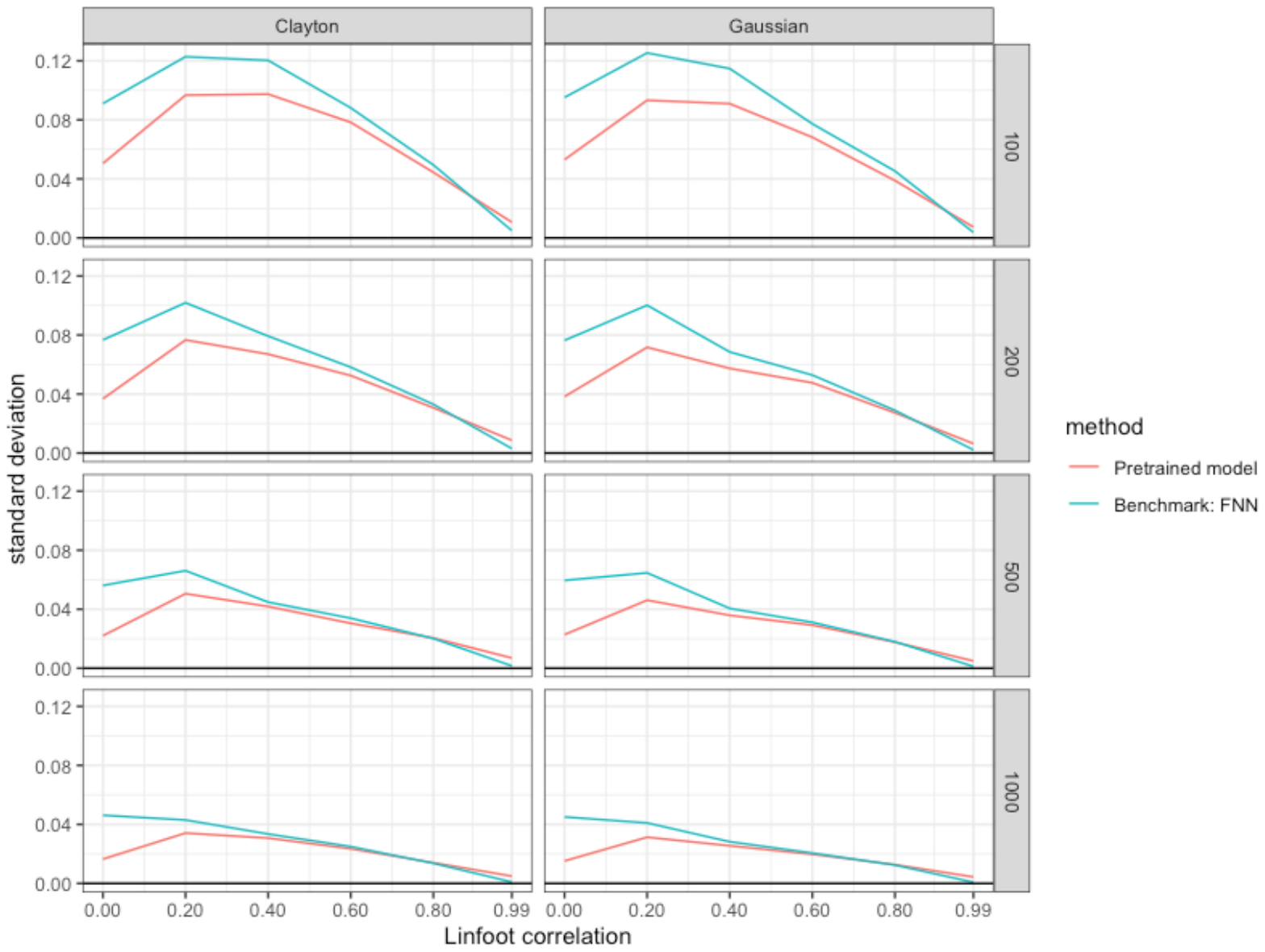}
\end{figure}

Figure 2 presents the standard deviation for the estimators. The pretrained neural network shows smaller variance than the FNN estimator, most notably for lower correlations.

\section{Application}\label{application}

We applied our pretrained estimator (Model 3) to BMI data on monozygotic (MZ) and dizygotic (DZ) twin pairs \cite{korkeila1991effects, hjelmborg2008genetic}. See Figure 3 for a scatter plot. Resemblance in twins is used to assess the genetic contribution to variation in phenotypes in
a given population. Under the equal environments assumption, more resemblance in MZ than in DZ twins is evidence of a heritable component. The Pearson correlations in MZ and DZ twins are sufficient statistics for the estimation of the heritability coefficient, but only in the realm of Gaussian distributions. Figure 3 clearly shows that the BMI data is not Gaussian.

\begin{figure}
\caption{Scatterplot of BMI for monozygotic (MZ) and dizygotic (DZ) twin pairs.}
\includegraphics[width=0.8\textwidth]{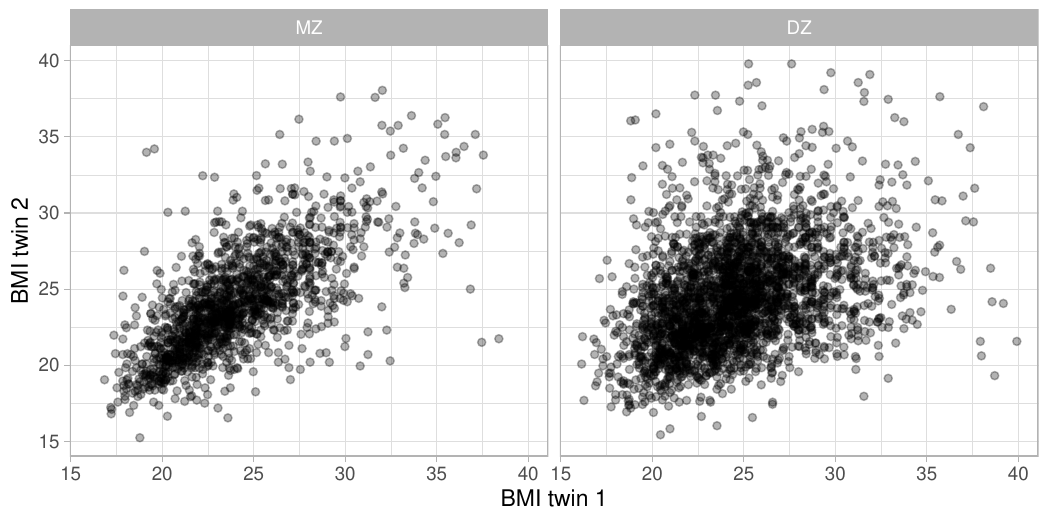}
\end{figure}

These data are a good motivation for a proper Linfoot correlation estimator to quantify the resemblance in MZ and DZ twins. Applying the Pearson correlation to twins
pairs with complete BMI data (\(n = 4,271\)), we found a correlation of \(0.68\) for the monozygotic (MZ) twins and a correlation of \(0.37\) for the dizygotic (DZ) twins. In contrast, our Linfoot estimator found a Linfoot correlation of \(0.74\) for the MZ twins and a Linfoot correlation of \(0.50\) for the DZ twins. From Figure 1 we saw that for sample sizes of 1000,
the bias of the estimator is rather small for the combined Linfoot estimator in the
range from \(0.20\) to \(0.80\). To find confidence intervals, we took 1000 bootstrap
samples from the original MZ and DZ datasets and computed the combined Linfoot estimator for
MZ and DZ. For MZ, the 95\% bootstrap confidence interval based on percentiles was \((0.70, 0.79)\) and for DZ the bootstrap interval was \((0.46, 0.54)\). Based on these results, we
can conclude that the MZ Linfoot correlation is significantly larger than the
DZ Linfoot correlation, suggesting a genetic contribution to the resemblance in BMI.

Although the Linfoot informational correlation will be invariant to the injective and continuous logarithmic transformation (via Theorem 2.1 \cite{beyond-normal-2023}), finite-sample estimates are not generally invariant. For our sample, a logarithmic transformation of the data changed the MZ Linfoot estimate from 0.74 to 0.72, and the DZ Linfoot estimate from 0.50 to 0.43.

\section{Discussion}\label{discussion}

The Pearson correlation is arguably the most favourite measure for dependency. It is appropriate
for Gaussian data. However, not all data are normal, and there are many alternatives for the Pearson correlation. A nice feature of the Linfoot informational correlation is that it has all the Rényi properties. Estimating the Linfoot correlation is however a challenge. We discussed several options when the distribution is unknown: non-parametric and neural estimators, and tracked their performance in a series of simulations of data from Gaussian and Clayton copulas. Overall, the performance of the non-parametric and neural estimators
was poor, with large bias and large variance. Among them, the FNN showed best performance in terms of bias and variance.

In the second part of this article we tried a different approach, inspired by deep (convolutional) neural networks. We benchmarked our neural network estimators with the FNN estimator. Overall best performance was shown by a deep convolutional neural network that fused output features based on handcrafted data summary statistics and image data containing a heatmap of the data. It showed overall low variance and low bias.

Our pretrained neural network estimator was not only better than the nonparametric methods, but also better than the MINE neural estimator. There are several key differences between neural estimators and our proposed estimator. Firstly, our network is pretrained on a large data set, whereas the current neural estimators are networks trained on the data for which the estimation is needed. This makes the currrent estimators slow and computer intensive, while our method, once it is trained, is very fast. We also observed a large repeatability variance for neural estimators: applying the estimator (estimation procedure) repeatedly on the same data set resulted in widely different estimates, whereas our pretrained estimator acts deterministically. On the other hand, neural estimators are universally applicable (in theory), whereas our pretrained model is only validated on data that are generated from Gaussian and Clayton copulas with Gaussian marginals, and only on certain sample sizes. For it to be more widely applicable it should receive extra training on more diverse types of data sets, with the limitation that for training, a ground truth is needed that is not always available.

In its current version, our Linfoot estimator can be used for any
bivariate data set. It is available on \texttt{github.com/pingapang/Pretrained-Linfoot-estimator}. If interest is on mutual information, it can easily be calculated from the Linfoot correlation. It should be noted that the estimator was trained on data with Gaussian margins. Since the Linfoot informational correlation is invariant to monotone transformations, a given data set without Gaussian margins can be transformed such that the marginals are close to Gaussian, and then fed into the deep neural network to obtain an estimate.

Performance of our method will however be highly dependent on the type of bivariate distribution. The model weights
are optimised to handle data that were used in the training. In the training,
we used particular sample sizes, and only used Clayton and Gaussian copulas, which are symmetric.
Moreover, in all simulated datasets, there was a positive relationship between
the two variables, where an increase in one variable was accompanied by an
increase in the other variable. Performance on data with negative relationships
was not tested but will undoubtedly be worse than on positive relationships. In future work,
the model should be retrained on more data sets so that it can deal with more varied situations.
Apart from focussing on negative relationship, future work should also look into
creating training data from other distributions than those generated by Clayton and Gaussian
copulas. The interesting feature of Linfoot correlations is that they are
applicable to any bivariate distribution and thus a more varied training
set should reflect the richness of possibilities. It is however an important
restriction of this machine learning method that for training we need ground-truth
labelling. For the Clayton copula we could find the true Linfoot correlation that
matches the Clayton copula parameter. Only for a limited set of distributions,
this relationship with the Linfoot correlation is known. To overcome this problem,
we can make use of the property that Linfoot correlations are invariant to
continuous injective transformations, as can be seen from applying the Linfoot transform to the Theorem 2.1 of \cite{beyond-normal-2023}. A more varied training set could therefore be generated
based on data generated by Archimedean copulas. For instance, homeomorphic transformations of the Clayton copula would yield invariant Linfoot values, by our Proposition \ref{prop:ClaytonLinfoot}. Further, our Proposition \ref{prop:linfootCopula} may yield closed formulas of the Linfoot correlation for other Archimedian copulas than the Clayton copula.

\section{Conclusion}\label{conclusion}

We studied various existing estimators of the Linfoot correlation, but found them to be
suboptimal. We showed that one promising direction is in the realm of deep learning.
A first version with a training set using a limited set of distributions showed promising results and
provides the field with a proof of principle that relatively unbiased estimators with low variance
can be constructed based on neural networks. Future work
will tell how far it can take us.

\section{Appendix}\label{appendix}

\begin{table}[htbp]
\center
{\begin{tabular}{l|c|c|c|c}
Sample size & $100$ & $200$ & $500$ & $1000$ \\
True Linfoot correlation & $0$ & $0$ & $0$ & $0$ \\
\hline
FNN
& 0.077 (0.098) & 0.060 (0.077) & 0.046 (0.058) & 0.040 (0.047) \\
KNN
& 0.154 (0.150) & 0.036 (0.076) & 0.012 (0.039) & \\
KNN, truncated
& 0.257 (0.067) & 0.220 (0.049) & 0.170 (0.031) & \\
KDE, MR
& 0.200 (0.061) & 0.204 (0.043) & 0.169 (0.028) & 0.149 (0.020) \\
KDE, beta
& 0.254 (0.042) & 0.233 (0.030) & 0.186 (0.020) & 0.161 (0.015) \\
MINE
& 0.188 (0.061)&0.138 (0.055)&0.067 (0.026)&0.049 (0.018)\\
\hline
Sample size & $100$ & $200$ & $500$ & $1000$ \\
True Linfoot correlation & $0.2$ & $0.2$ & $0.2$ & $0.2$ \\
\hline
FNN
& 0.155 (0.126) & 0.163 (0.098) & 0.184 (0.063) & 0.192 (0.039) \\
KNN
& 0.223 (0.165) & 0.116 (0.128) & 0.107 (0.100) & \\
KNN, truncated
& 0.310 (0.077) & 0.286 (0.057) & 0.257 (0.041) & \\
KDE, MR
& 0.258 (0.074) & 0.268 (0.051) & 0.248 (0.036) & 0.239 (0.026) \\
KDE, beta
& 0.298 (0.052) & 0.288 (0.043) & 0.260 (0.031) & 0.245 (0.024) \\
MINE
& 0.235 (0.066)& 0.193 (0.065)& 0.161 (0.035)&0.155 (0.025)\\
\hline
Sample size & $100$ & $200$ & $500$ & $1000$ \\
True Linfoot correlation & $0.4$ & $0.4$ & $0.4$ & $0.4$ \\
\hline
FNN
& 0.349 (0.113) & 0.373 (0.072) & 0.384 (0.042) & 0.392 (0.030) \\
KNN
& 0.416 (0.142) & 0.368 (0.114) & 0.387 (0.058) & \\
KNN, truncated
& 0.435 (0.084) & 0.431 (0.060) & 0.426 (0.039) & \\
KDE, MR
& 0.390 (0.073) & 0.411 (0.052) & 0.409 (0.035) & 0.406 (0.025) \\
KDE, beta
& 0.406 (0.059) & 0.416 (0.047) & 0.407 (0.032) & 0.407 (0.024) \\
MINE
&0.357 (0.077)&0.329 (0.077)&0.323 (0.034)&0.321 (0.024)\\
\hline
Sample size & $100$ & $200$ & $500$ & $1000$ \\
True Linfoot correlation & $0.6$ & $0.6$ & $0.6$ & $0.6$ \\
\hline
FNN
& 0.548 (0.077) & 0.570 (0.049) & 0.585 (0.032) & 0.592 (0.021) \\
KNN
& 0.625 (0.083) & 0.606 (0.059) & 0.612 (0.036) & \\
KNN, truncated
& 0.592 (0.065) & 0.599 (0.047) & 0.606 (0.030) & \\
KDE, MR
& 0.549 (0.062) & 0.577 (0.042) & 0.585 (0.028) & 0.588 (0.020) \\
KDE, beta
& 0.546 (0.055) & 0.567 (0.040) & 0.577 (0.026) & 0.586 (0.019) \\
MINE
&0.529 (0.070)&0.507 (0.072)&0.511 (0.030)&0.513 (0.021)\\
\hline
Sample size & $100$ & $200$ & $500$ & $1000$ \\
True Linfoot correlation & $0.8$ & $0.8$ & $0.8$ & $0.8$ \\
\hline
FNN
& 0.750 (0.045) & 0.768 (0.029) & 0.784 (0.018) & 0.792 (0.012) \\
KNN
& 0.815 (0.039) & 0.811 (0.030) & 0.816 (0.018) & \\
KNN, truncated
& 0.757 (0.041) & 0.774 (0.027) & 0.787 (0.017) & \\
KDE, MR
& 0.719 (0.036) & 0.752 (0.025) & 0.768 (0.016) & 0.777 (0.011) \\
KDE, beta
& 0.700 (0.032) & 0.735 (0.025) & 0.756 (0.016) & 0.771 (0.011) \\
MINE
&0.738 (0.047)&0.728 (0.046)&0.730 (0.021)&0.733 (0.015)\\
\hline
Sample size & $100$ & $200$ & $500$ & $1000$ \\
True Linfoot correlation & $0.99$ & $0.99$ & $0.99$ & $0.99$ \\
\hline
FNN
& 0.964 (0.004) & 0.974 (0.002) & 0.981 (0.001) & 0.985 (0.001) \\
KNN
& 0.985 (0.002) & 0.987 (0.002) & 0.990 (0.001) & \\
KNN, truncated
& 0.950 (0.005) & 0.964 (0.003) & 0.974 (0.001) & \\
KDE, MR
& 0.879 (0.003) & 0.923 (0.002) & 0.946 (0.001) & 0.958 (0.001) \\
KDE, beta
& 0.862 (0.002) & 0.901 (0.001) & 0.928 (0.001) & 0.945 (0.001) \\
MINE&0.977 (0.005)&0.974 (0.005)&0.971 (0.005)&0.976 (0.003)\\
\end{tabular}
\caption{Means and standard errors for the kernel, $k$-nearest neighbor and MINE estimators. Computations are based on Gaussian data with true Linfoot correlation $L \in \{0,0.2,0.4,0.6,0.8,0.99 \}$, $1000$ samples, and $100$, $200$, $500$, and $1000$ observations, respectively. There were convergence problems for the KNN estimators for sample sizes of 1000.}
}
\end{table}

\begin{table}[htbp]
\center
{\begin{tabular}{l|c|c|c|c}
Sample size & $100$ & $200$ & $500$ & $1000$ \\
True Linfoot correlation & $0$ & $0$ & $0$ & $0$ \\
\hline
FNN
& 0.075 (0.098) & 0.060 (0.076) & 0.044 (0.057) & 0.038 (0.046) \\
KNN
& 0.151 (0.155) & 0.038 (0.080) & 0.015 (0.044) & \\
KNN, truncated
& 0.256 (0.068) & 0.221 (0.047) & 0.171 (0.032) & \\
KDE, MR
& 0.197 (0.061) & 0.205 (0.041) & 0.168 (0.028) & 0.147 (0.019) \\
KDE, beta
& 0.252 (0.041) & 0.234 (0.029) & 0.186 (0.020) & 0.160 (0.014) \\
MINE
&0.185 (0.063)&0.139 (0.056)&0.065 (0.025)&0.051 (0.019)\\
\hline
Sample size & $100$ & $200$ & $500$ & $1000$ \\
True Linfoot correlation & $0.2$ & $0.2$ & $0.2$ & $0.2$ \\
\hline
FNN
& 0.139 (0.120) & 0.154 (0.102) & 0.165 (0.065) & 0.178 (0.043) \\
KNN
& 0.210 (0.167) & 0.092 (0.119) & 0.097 (0.099) & \\
KNN, truncated
& 0.298 (0.072) & 0.270 (0.057) & 0.242 (0.041) & \\
KDE, MR
& 0.250 (0.067) & 0.258 (0.051) & 0.239 (0.036) & 0.231 (0.027) \\
KDE, beta
& 0.291 (0.053) & 0.278 (0.041) & 0.252 (0.031) & 0.239 (0.024) \\
MINE
&0.225 (0.068)&0.184 (0.065)&0.143 (0.036)&0.138 (0.025)\\
\hline
Sample size & $100$ & $200$ & $500$ & $1000$ \\
True Linfoot correlation & $0.4$ & $0.4$ & $0.4$ & $0.4$ \\
\hline
FNN
& 0.321 (0.118) & 0.348 (0.076) & 0.369 (0.046) & 0.381 (0.033) \\
KNN
& 0.395 (0.155) & 0.345 (0.129) & 0.378 (0.066) & \\
KNN, truncated
& 0.400 (0.084) & 0.406 (0.062) & 0.400 (0.042) & \\
KDE, MR
& 0.369 (0.078) & 0.391 (0.056) & 0.390 (0.036) & 0.390 (0.027) \\
KDE, beta
& 0.388 (0.063) & 0.396 (0.050) & 0.392 (0.033) & 0.394 (0.026) \\
MINE
&0.330 (0.080)&0.297 (0.080)&0.288 (0.036)&0.285 (0.027)\\
\hline
Sample size & $100$ & $200$ & $500$ & $1000$ \\
True Linfoot correlation & $0.6$ & $0.6$ & $0.6$ & $0.6$ \\
\hline
FNN
& 0.521 (0.090) & 0.557 (0.058) & 0.576 (0.035) & 0.588 (0.024) \\
KNN
& 0.613 (0.094) & 0.597 (0.067) & 0.607 (0.039) & \\
KNN, truncated
& 0.560 (0.075) & 0.572 (0.053) & 0.584 (0.033) & \\
KDE, MR
& 0.525 (0.067) & 0.554 (0.048) & 0.564 (0.031) & 0.569 (0.022) \\
KDE, beta
& 0.524 (0.058) & 0.548 (0.044) & 0.559 (0.030) & 0.570 (0.022) \\
MINE
&0.482 (0.084)&0.459 (0.081)&0.466 (0.033)&0.468 (0.025)\\
\hline
Sample size & $100$ & $200$ & $500$ & $1000$ \\
True Linfoot correlation & $0.8$ & $0.8$ & $0.8$ & $0.8$ \\
\hline
FNN
& 0.735 (0.051) & 0.762 (0.033) & 0.781 (0.020) & 0.790 (0.014) \\
KNN
& 0.807 (0.045) & 0.806 (0.031) & 0.811 (0.019) & \\
KNN, truncated
& 0.738 (0.047) & 0.759 (0.032) & 0.775 (0.020) & \\
KDE, MR
& 0.696 (0.043) & 0.731 (0.029) & 0.749 (0.019) & 0.758 (0.013) \\
KDE, beta
& 0.680 (0.039) & 0.715 (0.027) & 0.738 (0.018) & 0.753 (0.013) \\
MINE
&0.692 (0.058)&0.681 (0.060)&0.683 (0.026)&0.684 (0.019)\\
\hline
Sample size & $100$ & $200$ & $500$ & $1000$ \\
True Linfoot correlation & $0.99$ & $0.99$ & $0.99$ & $0.99$ \\
\hline
FNN
& 0.960 (0.005) & 0.971 (0.003) & 0.979 (0.002) & 0.983 (0.001) \\
KNN
& 0.983 (0.003) & 0.986 (0.002) & 0.989 (0.001) & \\
KNN, truncated
& 0.948 (0.006) & 0.963 (0.003) & 0.973 (0.002) & \\
KDE, MR
& 0.877 (0.004) & 0.920 (0.003) & 0.942 (0.002) & 0.955 (0.001) \\
KDE, beta
& 0.856 (0.004) & 0.896 (0.003) & 0.922 (0.002) & 0.939 (0.001) \\
MINE
&0.964 (0.009)&0.961 (0.009)&0.958 (0.007)&0.964 (0.004)\\ \hline
\end{tabular}
\caption{Means and standard errors for the kernel, $k$-nearest neighbour and MINE estimators. Computations are based on Clayton data with true Linfoot correlation $L \in \{0,0.2,0.4,0.6,0.8,0.99 \}$, $1000$ samples, and $100$, $200$, $500$, and $1000$ observations, respectively. There were convergence problems for the KNN estimators for sample sizes of 1000.}
}
\end{table}

\appendix
\newpage

\section{Proof of proposition 1}

Let \((X,Y)\) be a pair of random variables taking values in \(\mathbb{R}^2\) with probability densities absolutely continuous with respect to the Lebesgue measure. The mutual information \(I(X,Y)\) between \(X\) and \(Y\) is defined as the quantity
\begin{align}
  I(X,Y) &= \E_{X,Y} \log \frac{f_{XY}(X,Y)}{f_X(X)f_Y(Y)} = \iint\limits_0^\infty f_{XY}(x,y)\log \frac{f_{XY}(x,y)}{f_X(x)f_Y(y)}dxdy\label{eq:mutualInformation}         
\end{align}
where \(f_{XY}\) is the joint density function of \((X,Y)\) and \(f_X\) and \(f_Y\) are the marginal density functions \cite{cover2012elements}.
\cite{Linfoot1957} introduced the informational correlation coefficient in as the one-to-one mapping \(l\colon\, \mathbb{R}_{\geq 0} \mapsto [0;1]\) of the mutual information
\begin{align}\label{eq:linfootDefinition}
  l\colon\, I(X, Y) \mapsto L(X,Y), \quad L(X,Y) = \sqrt{1 - \exp\left(-2I(X,Y)\right)}
\end{align}
where \(L(X,Y) = 0\) if and only if \(I(X,Y) = 0\) corresponding to independence and \(L(X,Y) \rightarrow 1\) for \(I(X,Y) \rightarrow \infty\) corresponding to strict dependence.

By Sklar's theorem \cite{sklar1959fonctions} the joint distribution \(F_{XY}\) can be written as \(F_{XY}(x,y) = C(F_{X}(x), F_Y(y))\) where \(F_{X}\) and \(F_{Y}\) are the marginal distribution functions and \(C\colon\, [0;1]^2 \mapsto [0;1]\) is the unique copula function modelling the dependence between the random pair. Applying the chain rule, the joint density function is given by
\begin{align}\label{eq:copuladensity}
  f_{XY}(x,y) = c(F_{X}(x), F_{Y}(y))f_{X}(x)f_{Y}(y)
\end{align}
where \(c(u,v) = \frac{\partial^2}{\partial u \partial v} C(u,v)\) is the copula density function. Combining the equations
\eqref{eq:mutualInformation} and \eqref{eq:copuladensity} and a change of variables we
obtain an expression of the mutual information as a functional of the copula density function.
\begin{align}
 I(X,Y) &= \iint\limits_{-\infty}^\infty \log \frac{f_{XY}(x,y)}{f_X(x)f_Y(y)} dF_{XY}(x,y) \nn\\ 
        &= \iint\limits_{-\infty}^\infty \log c(F_X(x), F_Y(y))dF_{XY}(x,y) \nn \\ 
        &= \iint\limits_0^1 \log c(u, v) dF_{XY}(F_X^{-1}(u), F_Y^{-1}(v)) \nn \\ 
        &= \iint\limits_0^1 \log c(u, v) dC(u, v) \nn \\ 
        &= \iint\limits_0^1 c(u,v) \log c(u,v) dudv \label{eq:mutualInfoViaCopuladensity}
\end{align}
Applying the Linfoot transformation from \eqref{eq:linfootDefinition} we obtain
\eqref{eq:linfootCopula}.

\section{Proof of proposition 2}

We consider Archimedean copulas with generator \(\psi: [0,\infty)\to [0,1]\), that is, \(\psi\) is a strictly convex decreasing function
with \(\psi(0)=1\). The corresponding Archimedean copula function satisfies \(C(u,v)=\psi(\psi^{-1}(u)+\psi^{-1}(v))\) for \((u,v)\in[0,1]^2\) and the copula density function is given by
\begin{align}\label{}
  c(u,v) &= \frac{\psi^{\prime\prime}\left(\psi^{-1}(u)+\psi^{-1}(v)\right)}{\psi^{\prime}\left(\psi^{-1}(u)\right)\psi^{\prime}\left(\psi^{-1}(v)\right)}
\end{align}
Now, for random variables \((X,Y) \in \mathbb{R}^2_+\)
such that \(X=\psi^{-1}(U)\) and \(Y=\psi^{-1}(V)\) the PDF is then given
by \(\psi^{\prime\prime}\left(x+y\right)\) on \(\mathbb{R}^2_+\) and hence
the mutual information in \eqref{eq:mutualInfoViaCopuladensity} can be expressed as
\begin{align}
I_\theta(X,Y) &= \iint\limits_0^\infty \psi_\theta^{\prime\prime}\left(x+y\right) \log \left(\frac{\psi_\theta^{\prime\prime}\left(x+y\right)}{\psi_\theta^{\prime}\left(x\right)\psi_\theta^{\prime}\left(y\right)} \right) dxdy\\
   &= (1+\theta) \log(1+\theta) I_1 + (1+\theta) \left(-2-\theta^{-1}\right) I_2 - 2(1+\theta)\left(-1-\theta^{-1}\right) I_3\nn
\end{align}
where \(\psi_\theta(t) = (1+\theta t)^{-1/\theta}\) is the inverse generator with first and second order derivatives \(\psi^{\prime}_\theta(t) =- (1+\theta t)^{-1-1/\theta }\) and \(\psi^{\prime\prime}_\theta(t) = (1+\theta)(1+\theta t)^{-2-1/\theta }\).
The integrals
\begin{align}
I_1 &= \iint\limits_0^\infty (1+\theta (x+y))^{-2-1/\theta } dxdy, \nn \\
I_2 &= \iint\limits_0^\infty (1+\theta (x+y))^{-2-1/\theta } \log \left(1+\theta (x+y)\right) dxdy, \nn \\
    I_3 &= \iint\limits_0^\infty (1+\theta (x+y))^{-2-1/\theta } \log\left(1+\theta y\right) dxdy \nn
\end{align}
are via standard calculations \(I_1 = (1+\theta)^{-1}\), \(I_2 = \theta (1+\theta)^{-1} + \theta (1+\theta)^{-2}\) and \(I_3 = \theta(1+\theta)^{-1}\) for $\theta>0$. Therefore
\begin{align}
 I_\theta(X,Y) &= \log(1+\theta)  - (2\theta+1)\left(1+\frac{1}{1+\theta}\right) + 2(1+\theta)
\end{align}
and thus using \eqref{eq:linfootDefinition}
\begin{align}
L_\theta(X,Y) &= \sqrt{1-\exp\left(-2\left(\log(1+\theta)  - (2\theta+1)\left(1+\frac{1}{1+\theta}\right) + 2(1+\theta)\right)\right)}
\end{align}
completes the proof.

\bibliography{Linfoot.bib}

\end{document}